\pdfoutput=1 
\documentclass[letterpaper]{article} 
\usepackage[]{aaai2027} 
\usepackage{times}  
\usepackage{helvet}  
\usepackage{courier}  
\usepackage[hyphens]{url}  
\usepackage{graphicx} 
\usepackage{natbib}  
\usepackage{caption} 
\usepackage{algorithm}
\usepackage{algorithmic}
\usepackage[table]{xcolor}
\usepackage{multirow}
\usepackage{amsmath}
\usepackage{amsfonts}
\usepackage{makecell}
\usepackage{tabularx}
\usepackage{booktabs}

\usepackage{newfloat}
\usepackage{listings}
\DeclareCaptionStyle{ruled}{labelfont=normalfont,labelsep=colon,strut=off} 
\floatstyle{ruled}
\newfloat{listing}{tb}{lst}{}
\floatname{listing}{Listing}
\title{Dual-Cache Latent Space Communication between \\ Heterogeneous Language Models}

\author{
    Jiyao Liu\textsuperscript{\rm 1}\equalcontrib,
    Qi Zhang\textsuperscript{\rm 2}\equalcontrib\footnote{Work does not relate to the author's position at Amazon.},
    Yaoyi Jia\textsuperscript{\rm 1},
    Ziwen Kan\textsuperscript{\rm 3},
    Song Wang\textsuperscript{\rm 3}
}
\affiliations{
    \textsuperscript{\rm 1}Independent Researcher\\
    \textsuperscript{\rm 2}Amazon Web Services (AWS)\\
    \textsuperscript{\rm 3}University of Central Florida (UCF)\\
    jiyao.liu@outlook.com,
    qizmax@amazon.com,
    yaoyi.jia@colorado.edu,
    song.wang@ucf.edu
}

\newcommand{\xkv}{\textsc{XKV}}
\newcommand{\lcfx}{\textsc{LCF-X}}        
\newcommand{\ttt}{\textsc{T2T}}           
\newcommand{\sharer}{\textsf{Sharer}}
\newcommand{\receiver}{\textsf{Receiver}}

\begin{document}

\maketitle

\begin{abstract}
Multi-agent LLM systems split work across models, so answering often
requires knowledge that sits in another agent's context: a \sharer{} has
encoded information that a \receiver{} needs to complete its task. Such
agents usually communicate by exchanging text, which places autoregressive
decoding on the critical path and reduces the exchange to a discrete message
written without sight of the receiver's state. Recent latent protocols
instead translate the sharer's key--value (KV) cache into the receiver's:
C2C supports heterogeneous models but requires both to read the same input,
while LCF-X removes this shared-context requirement through position-free
sharer-cache pooling. Three restrictions remain: LCF-X compresses the
sharer alone, supplies the same layer-local summary to every receiver
position with no joint cross-layer memory to retrieve from, and assumes
matched layer count and KV geometry.
We introduce \xkv{}, which lifts all three with three components:
learned-query attention pools \emph{both} caches; self-attention over
receiver-aligned layer tokens, with a learned layer map reconciling
different depths, mixes the pooled summaries into a compact joint memory;
and a shared position decoder lets every raw receiver cache position
retrieve its own per-head-gated residual in the receiver's native KV
geometry. Both models stay frozen and may differ in family, depth, KV-head
count, head dimension, and tokenizer; only the translator is trained.
Across 45 dataset--model-pair settings (six heterogeneous and three
same-model ordered pairings, five datasets), \xkv{} attains the highest
macro score and best average rank of the three protocols, improving on
LCF-X on every dataset (by 4.6 exact-match and 4.2 F1 points on ROPES) and
surpassing text communication on four of the five, while training 76\%
fewer parameters and translating a cache pair $10.3\times$ faster
($5.8$ against $59.9$\,ms); end to end, \xkv{} is 26\% faster than LCF-X
and $6.8\times$ faster than text communication.
\end{abstract}

\section{Introduction}

Language models increasingly act as components of larger systems, retrieving
evidence, invoking tools, and verifying one another
\citep{yao2023react,du2023debate,wu2023autogen,chen2024agentverse}. In such
systems, agents routinely observe different things: retrieved documents are
partitioned across agents, tools return results only to their caller, and
each interaction leaves private state behind. Whenever one agent needs
evidence another has already encoded, the two must communicate, and
collective performance depends on that channel. We study its two-model
core, the \emph{cross-context} setting: a \sharer{} has encoded part of the
evidence for a question, a \receiver{} has encoded the remainder and must
produce the answer; neither context suffices alone, so the receiver must
integrate the evidence it lacks with what it already holds.
Deployed systems address this almost exclusively by exchanging natural
language. Text is portable and interpretable, but between neural models it is
indirect: the sender autoregressively decodes its continuous state into
tokens on the critical path, and the receiver must tokenize and re-encode
them. Cache-to-Cache (C2C) eliminates this decode--encode cycle by
translating one model's key--value (KV) cache into another's
\citep{fu2026cache,dery2026latent}. It supports heterogeneous families and
tokenizers, but aligns the two tokenizations of the same underlying input
and fuses semantically corresponding positions, so it remains a
shared-context method; cache-reuse methods that realign positions make a
related correspondence assumption \citep{ye2025kvcomm}. The cross-context
extension of Latent Cache Flow, \lcfx{}, removes this presupposition with a
pooled, position-free cache summary \citep{rossi2026latentcacheflow},
showing that evidence can move at the cache level even when the contexts
share no tokens.

\begin{figure}[t]
  \centering
  \resizebox{\columnwidth}{!}{%
    \begin{picture}(240,230)
      \put(0,0){\includegraphics[width=240pt]{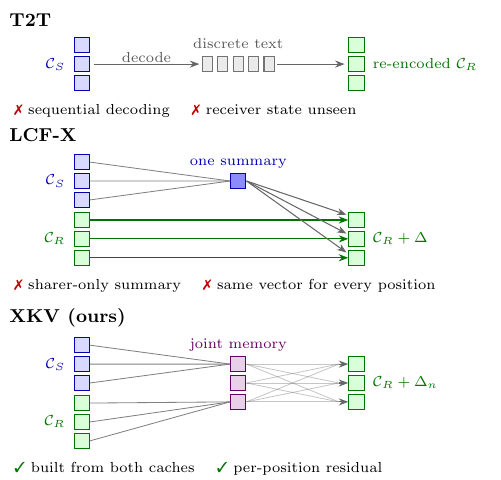}}
      \put(90,84){\color{white}\rule{150pt}{18pt}}
      \put(92,88){\scriptsize
        $\times$ one shared summary per layer}
    \end{picture}}
  \vspace{-17pt}
  \caption{Communication protocols for two models holding complementary
  contexts, annotated with the challenges each raises
  (\textcolor{red}{$\times$}) or resolves
  (\textcolor{green!45!black}{$\checkmark$}). \ttt{} decodes a message on the
  critical path without access to the receiver's state. \lcfx{} pools a
  summary from the sharer cache alone and supplies the same summary to
  every position, although its downstream projector also conditions on the
  local receiver cache row. \xkv{} (ours) builds the communication memory jointly from both
  caches, and each receiver position retrieves its own receiver-native
  residual.}
  \label{fig:protocols}
  \vspace{-17pt}
\end{figure}

Position-independent transfer, however, leaves three restrictions in
place, contrasted in Figure~\ref{fig:protocols}. \textbf{First, what to
communicate is decided without reference to the receiver}: only the sharer
cache is pooled (\lcfx{}'s projector reads the receiver cache only after
the summary is formed), so one vector must carry every fact the receiver
might need, selected without consulting what it has already encoded.
\textbf{Second, the external signal is a layer-local, single-summary
bottleneck}: every position of a receiver layer gets the same pooled sharer
summary; the projector's view of the local cache row lets residuals vary
across positions, but no position can select among multiple sharer
summaries or retrieve information from other layers.
\textbf{Third, flexibility is limited}: C2C accepts heterogeneous models
but requires shared, token-alignable content, and original \lcfx{} accepts
different content but assumes matched layer count and KV-head geometry; a
practical channel should accept a frozen pair that differs along these
axes. All three must be addressed under an efficiency constraint: a
receiver-aware channel that reinstates the cost latent communication was
introduced to remove yields no net benefit.

\begin{figure}[t]
  \centering
  \includegraphics[width=\columnwidth]{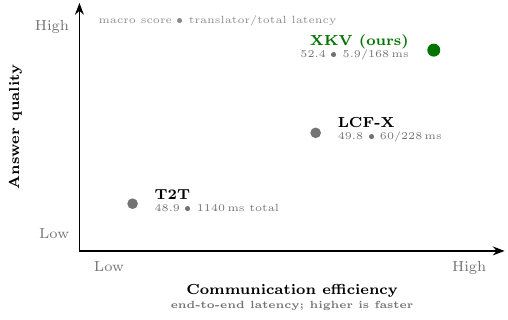}
  \vspace{-17pt}
  \caption{Answer quality versus communication cost, aggregated over all 45
  dataset--model-pair settings (details in the technical appendix); the
  ordering on both axes is faithful, the spacing is not to scale. \xkv{}
  dominates both baselines: it improves over \lcfx{} on every dataset with a
  $10.3\times$ faster translator and 76\% fewer trainable parameters, and is
  $6.8\times$ faster end to end than \ttt{}, which leads only on QASC. C2C is
  omitted: it requires token-aligned shared context.}
  \label{fig:positioning}
  \vspace{-17pt}
\end{figure}

We resolve these with \xkv{}, a dual-cache translator that separates \emph{what
to communicate} from \emph{where to write it}. Learned-query attention pools
\emph{both} caches into $k$ candidate summaries per layer and head. Since
depth carries no reliable semantic correspondence across architectures
\citep{kornblith2019similarity,bansal2021stitching}, a learned layer map
reconciles the two depth axes, and self-attention mixes the summaries into
a compact joint memory that represents the sharer's evidence relative to
what the receiver has encoded, not as a fixed sender-authored message.
Each raw receiver cache position then queries this memory through a shared
decoder, and zero-initialized output heads emit per-head-gated residuals in
the receiver's native cache geometry, giving every position a distinct
update. Efficiency follows from the same choices rather than trading
against them: pooling reduces $N\!\rightarrow\!k$ in one vectorized step
rather than two hard-to-batch stages over variable-length spans; the
memory has length $L_R$ regardless of prompt length; and the decoder,
gate, and output heads are shared across receiver layers. Both models
remain frozen and may differ in depth, KV-head count, head dimension, and
tokenizer; only the translator is trained.

We evaluate \xkv{} on five split-evidence reasoning tasks
\citep{yang2018hotpotqa,trivedi2022musique,khot2020qasc,lin2019ropes,geva2021strategyqa}
over the complete $3\!\times\!3$ grid of Qwen, Gemma, and Llama
sharer--receiver pairs
\citep{yang2025qwen3,gemmateam2025gemma3,grattafiori2024llama3}, whose six
off-diagonal pairs collectively span family, depth, attention-geometry, and
tokenizer differences; since original \lcfx{} does not support such
mismatches, we extend it into a stronger heterogeneous baseline.
Across all 45 dataset--model-pair settings, \xkv{} attains the
highest macro score (52.4 against 49.8 for \lcfx{} and 48.9 for \ttt{}),
improving on \lcfx{} on every dataset, while its translator runs
$10.3\times$ faster than the \lcfx{} fusor with 76\% fewer trainable
parameters; end to end it is 26\% faster than \lcfx{} and $6.8\times$
faster than \ttt{} (Figure~\ref{fig:positioning}). The contribution is
therefore the removal of a trade-off: communication that is receiver-aware
and retrieves position-specific updates from a joint cross-layer memory,
yet is strictly cheaper than the sender-centric summary it replaces.

We make three contributions:
\begin{itemize}
  \item We identify receiver-independent compression, layer-local
  single-summary conditioning, and matched-geometry assumptions as
  bottlenecks of cross-context cache communication, and reformulate the
  task as efficient joint translation of both models' states.
  \item We introduce \xkv{}, combining symmetric direct cache pooling,
  cross-layer translation, and receiver-position queries into a joint
  memory for receiver-native reconstruction, with both frozen models free
  to differ in architecture and tokenizer.
  \item We extend \lcfx{} to mismatched layer counts and KV dimensions and
  conduct, to our knowledge, the first full cross-family, split-evidence
  evaluation (five tasks, nine ordered pairs), where \xkv{} improves on
  both the strengthened \lcfx{} baseline and text communication at a
  fraction of their cost.
\end{itemize}


\section{Related Work}
\paragraph{Communication between language-model agents.}
Deployed multi-agent systems coordinate exclusively through natural
language: agents exchange textual arguments in debate \citep{du2023debate},
converse in AutoGen \citep{wu2023autogen}, and collaborate in
role-specialized teams \citep{chen2024agentverse}, paying autoregressive
generation and re-encoding at every handoff; our \ttt{} baseline follows this
protocol. 
A second line avoids committing intermediate computation to
language: Coconut feeds hidden states back as continuous reasoning
\citep{hao2024coconut}, Interlat learns compressed hidden-state communication
between agents \citep{du2026interlat}, and LatentMAS maintains a shared
latent working memory \citep{zou2025latentmas}. 
These methods operate on
hidden-state sequences or a shared latent interface; \xkv{} instead targets
frozen, independently tokenized models and writes into the per-layer KV state
that the receiver's attention already consumes.

\paragraph{KV caches as a communication medium.}
The KV cache is the persistent prompt state consumed during decoding
\citep{pope2023efficiently}. 
Within one model, eviction policies
\citep{zhang2023h2o,li2024snapkv,xiao2024streamingllm} and latent projections
\citep{deepseekai2024v2} show that state is highly redundant, which is
what makes a compact cross-model channel plausible. Between models, C2C
learns adapters that support heterogeneous models while fusing
position-aligned caches of the same underlying input
\citep{fu2026cache,dery2026latent}, and KV entries can be selectively shared
\citep{shi2026kvcomm} or reused across agents after positional realignment
\citep{ye2025kvcomm}; all assume overlapping content, but our two models
hold complementary evidence. 
Most closely, \lcfx{} pools the sharer cache
into a position-free summary before layer-local residual fusion
\citep{rossi2026latentcacheflow}. 
\xkv{} advances this line by building a
cross-layer memory from both caches and letting each receiver position
retrieve a distinct update through shared parameters.

\paragraph{Bridging frozen representation spaces.}
Translating between frozen models presumes their states are relatable:
similarity analyses find broadly comparable features that resist naive
layer-to-layer identification \citep{kornblith2019similarity}, stitching
shows a light learned map often suffices \citep{bansal2021stitching}, and
relative representations communicate zero-shot through shared anchors
\citep{moschella2023relative}; together these motivate our learned layer map
and learned translation. Prefix tuning \citep{li2021prefixtuning} and gist
tokens \citep{mu2023gist} likewise inject continuous vectors into a frozen
model, but as a static prefix or single-model compression rather than an
example-dependent translation of two models' states. Finally, pooling with
learned queries is a standard primitive
\citep{lee2019settransformer,jaegle2021perceiver}; \xkv{} applies it per
layer and KV head to both caches at once, so the resulting slots stay
cache-shaped and can be reassembled into receiver-native keys and values.

\begin{figure}[t]
  \centering
  \includegraphics[width=\columnwidth]{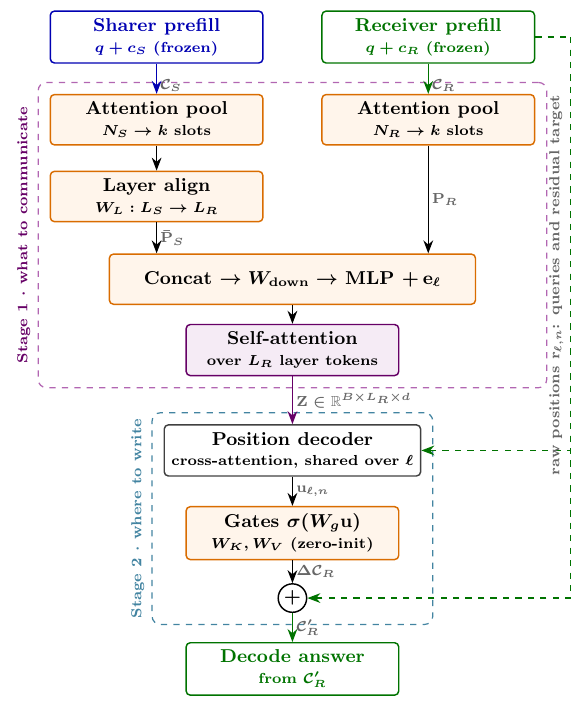}
  \vspace{-15pt}
  \caption{Overview of \xkv. Both frozen models prefill their private
  contexts. Stage~1 (\emph{what to communicate}): learned-query attention
  pools each cache from $N_a$ positions to $k$ slots per layer and head, a
  learned map aligns the sharer layer axis, and the concatenated summaries are
  bottlenecked and mixed by self-attention into one joint memory token per
  receiver layer. Stage~2 (\emph{where to write}): every raw receiver cache
  position queries the memory through a shared cross-attention decoder, and
  shared, zero-initialized output heads emit per-head-gated KV residuals in
  the receiver's native cache geometry. Only the \xkv{} translator is trained.}
  \label{fig:xkv_architecture}
  \vspace{-15pt}
\end{figure}

\section{Method}

\xkv{} is a lightweight translator between two frozen language models,
built on a single principle: \emph{what to communicate} should be decided
jointly from both models' encoded states, whereas \emph{where to write it}
should be dictated by the receiver's own cache geometry. Accordingly,
\xkv{} first compresses both KV caches into a compact joint layer memory
through symmetric pooling, layer alignment, and cross-layer mixing, then
lets every raw receiver cache position retrieve its own update from that
memory, written back as receiver-native KV residuals
(Figure~\ref{fig:xkv_architecture}).

\subsection{Problem Setup}

Let a frozen \sharer{} model $S$ observe a question $q$ and private context
$c_S$, and let a frozen \receiver{} model $R$ observe $q$ and complementary
context $c_R$. The receiver must predict answer tokens
$y=(y_1,\ldots,y_T)$. Prefilling model $a\in\{S,R\}$ produces a KV cache
\begin{equation}
\mathcal{C}_a =
\left\{(\mathbf{K}^{\ell}_a,\mathbf{V}^{\ell}_a)\right\}_{\ell=1}^{L_a},
\quad
\mathbf{K}^{\ell}_a,\mathbf{V}^{\ell}_a
\in\mathbb{R}^{B\times H_a\times N_a\times D_a},
\label{eq:cache}
\end{equation}
where $L_a$, $H_a$, $N_a$, and $D_a$ are the number of layers, KV heads,
cached positions, and the head dimension. These quantities and the tokenizer
may differ between $S$ and $R$; architectures that interleave full and
sliding-window attention are handled by operating on the common valid suffix
of the per-layer caches.
\xkv{} is a learned translator
$F_\theta(\mathcal{C}_S,\mathcal{C}_R)=\Delta\mathcal{C}_R$
that produces receiver-shaped residuals and updates the prompt cache once,
\begin{equation}
\mathcal{C}'_R = \mathcal{C}_R+\Delta\mathcal{C}_R,
\label{eq:update}
\end{equation}
after which the receiver generates from $\mathcal{C}'_R$ with its standard
causal decoder. Both language models stay frozen; only $\theta$ is trained.

\subsection{Symmetric Direct Cache Pooling}

\paragraph{Motivation.}
A single sharer-only summary forces several useful source facts to compete for
one bottleneck and cannot express their relevance to what the receiver has
already encoded; we therefore keep $k$ candidate summaries per layer and head,
and pool \emph{both} caches with the same mechanism.

For side $a\in\{S,R\}$, layer $\ell$, KV head $h$, and slot
$j\in\{1,\ldots,k\}$, \xkv{} learns a query
$\mathbf{q}_{a,\ell,h,j}\in\mathbb{R}^{D_a}$. Keys determine attention weights
over the raw cache positions:
\begin{equation}
A_{a,\ell,h,j,n}
=
\operatorname{softmax}_{n}
\left(
\frac{\mathbf{q}_{a,\ell,h,j}^{\top}
\mathbf{K}_{a,\ell,h,n}}{\sqrt{D_a}}
\right),
\label{eq:pool-attn}
\end{equation}
and the same weights pool keys and values,
\begin{equation}
\mathbf{p}^{X}_{a,\ell,h,j}
=\sum_{n=1}^{N_a} A_{a,\ell,h,j,n}\mathbf{X}_{a,\ell,h,n},
\quad
\mathbf{X}\in\{\mathbf{K},\mathbf{V}\},
\label{eq:pool}
\end{equation}
yielding
$\mathbf{P}^{K}_a,\mathbf{P}^{V}_a
\in\mathbb{R}^{B\times L_a\times H_a\times k\times D_a}$.
Padding positions are masked out of the softmax. Unlike \lcfx's hierarchical
within-span and then across-span pooling \citep{rossi2026latentcacheflow},
this operation reduces
$N_a\!\rightarrow\!k$ in one step and is fully vectorized across layers,
heads, and slots. In the default configuration, $k\in\{1,2,3\}$ grows with
receiver model size.

\subsection{Layer Alignment and Joint Memory}

\paragraph{Motivation.}
Depth provides no reliable one-to-one semantic correspondence across
architectures \citep{kornblith2019similarity}, so the memory should be
receiver-aligned yet allow both depth conversion and communication among
layers.

When $L_S\ne L_R$, an affine map along the layer axis,
$\bar{\mathbf{P}}_S=W_L\mathbf{P}_S+\mathbf{b}_L$ with
$W_L\in\mathbb{R}^{L_R\times L_S}$ applied independently to every pooled
feature, converts the sharer summaries to $L_R$ layers; the identity is used
when depths match. For receiver layer $\ell$, all $k$ pooled slots of both
sides are concatenated into one feature vector,
\begin{equation}
\mathbf{x}_{\ell}
=
\left[
\operatorname{vec}(\bar{\mathbf{P}}^{K}_{S,\ell});
\operatorname{vec}(\bar{\mathbf{P}}^{V}_{S,\ell});
\operatorname{vec}(\mathbf{P}^{K}_{R,\ell});
\operatorname{vec}(\mathbf{P}^{V}_{R,\ell})
\right].
\label{eq:concat}
\end{equation}
A shared down-projection and bottleneck MLP produce one memory token per
receiver layer,
\begin{equation}
\mathbf{z}^{0}_{\ell}
=
\operatorname{LN}\!\left(
\operatorname{MLP}(W_{\mathrm{down}}\mathbf{x}_{\ell})
\right)+\mathbf{e}_{\ell},
\quad
\mathbf{Z}^{0}\in\mathbb{R}^{B\times L_R\times d},
\label{eq:bottleneck}
\end{equation}
where $\mathbf{e}_{\ell}$ is a learned receiver-layer embedding and the latent
width $d$ defaults to one eighth of the receiver hidden width. A
pre-normalized Transformer block then mixes information across the
receiver-aligned layer tokens:
\begin{align}
\tilde{\mathbf{Z}}
&=\mathbf{Z}^{0}+
\operatorname{MHA}(\operatorname{LN}(\mathbf{Z}^{0})),
\label{eq:mix-attn}\\
\mathbf{Z}
&=\tilde{\mathbf{Z}}+
\operatorname{FFN}(\operatorname{LN}(\tilde{\mathbf{Z}})).
\label{eq:mix-ffn}
\end{align}
The resulting $\mathbf{Z}$ is a joint, cross-layer communication memory. Its
sequence length is $L_R$, independent of either prompt length, and it
represents the sharer's evidence \emph{in relation to} what the receiver has
already encoded rather than as a fixed sender-authored message.

\subsection{Receiver-Position Retrieval and Update}

\paragraph{Motivation.}
Global pooling removes token locality. In LCF-X, every position receives
the same external sharer summary, although a downstream projector also reads
the local receiver cache and may therefore emit position-varying residuals.
XKV instead uses the original receiver cache entries as explicit queries into
a multi-slot, cross-layer memory, so each position can retrieve a different
combination of sharer and receiver information.

For layer $\ell$ and position $n$, the raw receiver key and value are
flattened across heads and concatenated,
\begin{equation}
\mathbf{r}_{\ell,n}
=
\left[
\operatorname{vec}(\mathbf{K}^{\ell}_{R,:,n,:});
\operatorname{vec}(\mathbf{V}^{\ell}_{R,:,n,:})
\right],
\label{eq:rows}
\end{equation}
and a decoder shared across all receiver layers forms queries
$\mathbf{q}_{\ell,n}=W_q\mathbf{r}_{\ell,n}+\mathbf{e}_{\ell}$, which
cross-attend to all $L_R$ memory tokens:
\begin{align}
\tilde{\mathbf{u}}_{\ell}
&=\mathbf{q}_{\ell}+
\operatorname{MHA}\!\left(
\operatorname{LN}(\mathbf{q}_{\ell}),\mathbf{Z},\mathbf{Z}
\right),
\label{eq:retrieve-attn}\\
\mathbf{u}_{\ell}
&=\tilde{\mathbf{u}}_{\ell}
+\operatorname{FFN}(\operatorname{LN}(\tilde{\mathbf{u}}_{\ell})).
\label{eq:retrieve-ffn}
\end{align}
The layer embedding $\mathbf{e}_{\ell}$ preserves layer identity under weight
sharing. Shared output heads then predict receiver-shaped updates, with a
deterministic sigmoid gate providing separate scales for every position, KV
head, and cache type:
\begin{equation}
[\boldsymbol{\alpha}^{K}_{\ell,n};
 \boldsymbol{\alpha}^{V}_{\ell,n}]
=\sigma(W_g\mathbf{u}_{\ell,n}),
\label{eq:gate}
\end{equation}
\begin{align}
\Delta\mathbf{K}^{\ell}_{R,:,n,:}
&=\boldsymbol{\alpha}^{K}_{\ell,n}\odot
\operatorname{reshape}(W_K\mathbf{u}_{\ell,n}),
\label{eq:res-k}\\
\Delta\mathbf{V}^{\ell}_{R,:,n,:}
&=\boldsymbol{\alpha}^{V}_{\ell,n}\odot
\operatorname{reshape}(W_V\mathbf{u}_{\ell,n}).
\label{eq:res-v}
\end{align}
Residuals at padding positions are zeroed, and the remaining residuals are
added to the receiver cache as in Eq.~\eqref{eq:update}.

\paragraph{Identity at initialization.}
$W_K$ and $W_V$ are shared across receiver layers and initialized to zero, so
an untrained translator reproduces receiver-only decoding exactly; training
therefore learns a residual \emph{correction} of the receiver cache rather
than a replacement, which stabilizes early optimization.

\subsection{Training and Inference}

Both base models are frozen; only $\theta$ is trained with teacher-forced
cross-entropy on the answer tokens:
\begin{equation}
\mathcal{L}(\theta)
=-\frac{1}{T}\sum_{t=1}^{T}
\log p_R\!\left(
y_t\mid y_{<t},q,c_R,
F_\theta(\mathcal{C}_S,\mathcal{C}_R)
\right).
\label{eq:loss}
\end{equation}
Although the receiver weights are fixed, gradients flow through its attention
computation into the \xkv{} residuals. At inference, the receiver prefills all
but the final prompt token, \xkv{} modifies that prefix cache once, and the
held-out token starts ordinary greedy decoding. Cache entries created for
generated tokens are appended normally and never retroactively rewritten, so
the one-shot update adds no per-token decoding cost.

\begingroup
\let\pairwisetextbf\textbf
\renewcommand{\textbf}[1]{\underline{\pairwisetextbf{#1}}}
\begin{table*}[!t]
  \centering
  \small
  \renewcommand{\arraystretch}{.99}
  \setlength{\tabcolsep}{6.4pt}
  \caption{Full results for all nine ordered sharer--receiver settings. Each
  setting occupies three rows, one per communication method. We report EM/F1
  for generative QA and accuracy (Acc.) for QASC and StrategyQA; higher is
  better. Cross-dataset Avg. averages the primary metric (F1 for generative
  QA and accuracy for classification) across all five datasets. The technical
  appendix provides the same results as per-dataset tables with cell-wise
  highlighting.}
  \vspace{-5pt}
  \label{tab:full-pairwise-main}
  \begin{tabular}{llccccccccc}
    \toprule
    \multirow{2}{*}{Sharer $\rightarrow$ Receiver} & \multirow{2}{*}{Method}
      & \multicolumn{2}{c}{ROPES} & \multicolumn{2}{c}{MuSiQue}
      & \multicolumn{1}{c}{QASC} & \multicolumn{1}{c}{StrategyQA}
      & \multicolumn{2}{c}{HotpotQA-bridge}
      & \multirow{2}{*}{\makecell{Cross-dataset\\Avg.}} \\
    \cmidrule(lr){3-4}\cmidrule(lr){5-6}\cmidrule(lr){7-7}\cmidrule(lr){8-8}\cmidrule(lr){9-10}
      & & EM & F1 & EM & F1 & Acc. & Acc. & EM & F1 & \\
    \midrule
    \rowcolor{blue!6} & \textsc{T2T}   & 35.8 & 45.2 & 4.6 & 12.2 & 78.6 & \textbf{64.1} & 16.6 & 27.1 & 45.44 \\
    \rowcolor{blue!6} & \textsc{LCF-X} & 45.2 & 52.4 & 7.7 & 15.4 & \textbf{82.4} & 58.8 & 29.1 & 40.6 & 49.92 \\
    \rowcolor{blue!6}\multirow{-3}{*}{Qwen $\rightarrow$ Qwen} & \textsc{XKV}   & \textbf{47.9} & \textbf{55.6} & \textbf{8.1} & \textbf{16.0} & 82.2 & 61.4 & \textbf{29.5} & \textbf{40.7} & \textbf{51.18} \\
    \multirow{3}{*}{Qwen $\rightarrow$ Gemma}
      & \textsc{T2T}   & 42.4 & 48.6 & \textbf{6.8} & \textbf{14.5} & \textbf{65.8} & 51.9 & 22.4 & 34.0 & \textbf{42.96} \\
      & \textsc{LCF-X} & 43.9 & 51.3 & 6.0 & 13.3 & 49.2 & \textbf{60.8} & 23.3 & 34.0 & 41.72 \\
      & \textsc{XKV}   & \textbf{46.2} & \textbf{54.3} & 5.4 & 12.5 & 55.1 & 58.2 & \textbf{23.8} & \textbf{34.7} & \textbf{42.96} \\
    \rowcolor{blue!6} & \textsc{T2T}   & 50.6 & 58.0 & 9.3 & 19.5 & 86.7 & 73.4 & 30.4 & 44.0 & 56.32 \\
    \rowcolor{blue!6} & \textsc{LCF-X} & 51.4 & 58.5 & 16.0 & 26.3 & 92.0 & 72.9 & 41.1 & 54.5 & 60.84 \\
    \rowcolor{blue!6}\multirow{-3}{*}{Qwen $\rightarrow$ Llama} & \textsc{XKV}   & \textbf{57.3} & \textbf{61.8} & \textbf{19.2} & \textbf{29.9} & \textbf{93.3} & \textbf{73.6} & \textbf{41.5} & \textbf{55.0} & \textbf{62.72} \\
    \multirow{3}{*}{Gemma $\rightarrow$ Qwen}
      & \textsc{T2T}   & 31.9 & 42.5 & 3.4 & 10.8 & \textbf{84.6} & \textbf{65.9} & 15.5 & 25.7 & 45.90 \\
      & \textsc{LCF-X} & 47.2 & 53.7 & \textbf{8.4} & \textbf{16.6} & 81.3 & 59.3 & 29.4 & 40.9 & 50.36 \\
      & \textsc{XKV}   & \textbf{49.1} & \textbf{55.4} & 8.1 & 15.8 & 82.4 & 62.0 & \textbf{29.5} & \textbf{41.0} & \textbf{51.32} \\
    \rowcolor{blue!6} & \textsc{T2T}   & 31.8 & 42.1 & \textbf{6.2} & \textbf{13.4} & \textbf{55.7} & 57.6 & 21.7 & 33.1 & 40.38 \\
    \rowcolor{blue!6} & \textsc{LCF-X} & 45.5 & 51.4 & 4.6 & 11.9 & 44.8 & 58.2 & 23.5 & \textbf{34.4} & 40.14 \\
    \rowcolor{blue!6}\multirow{-3}{*}{Gemma $\rightarrow$ Gemma} & \textsc{XKV}   & \textbf{48.1} & \textbf{55.6} & 5.8 & 13.1 & 53.9 & \textbf{58.8} & \textbf{23.9} & \textbf{34.4} & \textbf{43.16} \\
    \multirow{3}{*}{Gemma $\rightarrow$ Llama}
      & \textsc{T2T}   & 46.9 & 55.7 & 9.4 & 18.3 & 89.4 & 70.9 & 30.4 & 44.5 & 55.76 \\
      & \textsc{LCF-X} & 42.1 & 49.8 & 17.8 & 28.3 & 92.9 & 68.8 & 40.2 & 53.5 & 58.66 \\
      & \textsc{XKV}   & \textbf{54.0} & \textbf{58.7} & \textbf{19.4} & \textbf{30.0} & \textbf{93.0} & \textbf{74.1} & \textbf{41.5} & \textbf{55.1} & \textbf{62.18} \\
    \rowcolor{blue!6} & \textsc{T2T}   & 38.6 & 47.7 & 5.9 & 15.4 & \textbf{85.0} & 65.6 & 21.1 & 32.9 & 49.32 \\
    \rowcolor{blue!6} & \textsc{LCF-X} & 46.7 & 54.8 & 7.6 & 15.5 & 81.1 & 53.1 & 28.6 & 40.3 & 48.96 \\
    \rowcolor{blue!6}\multirow{-3}{*}{Llama $\rightarrow$ Qwen} & \textsc{XKV}   & \textbf{53.0} & \textbf{59.0} & \textbf{8.5} & \textbf{16.5} & 82.1 & \textbf{65.8} & \textbf{29.3} & \textbf{40.9} & \textbf{52.86} \\
    \multirow{3}{*}{Llama $\rightarrow$ Gemma}
      & \textsc{T2T}   & 40.3 & 48.3 & \textbf{12.1} & \textbf{21.9} & \textbf{55.9} & 53.7 & \textbf{28.3} & \textbf{41.5} & \textbf{44.26} \\
      & \textsc{LCF-X} & 45.1 & 50.4 & 5.4 & 12.4 & 44.9 & 52.2 & 23.5 & 34.4 & 38.86 \\
      & \textsc{XKV}   & \textbf{48.3} & \textbf{55.7} & 5.2 & 12.2 & 54.9 & \textbf{57.3} & 24.4 & 35.0 & 43.02 \\
    \rowcolor{blue!6} & \textsc{T2T}   & 48.4 & 55.7 & 14.2 & 26.3 & 90.7 & 72.1 & 36.1 & 52.0 & 59.36 \\
    \rowcolor{blue!6} & \textsc{LCF-X} & 45.6 & 52.8 & 16.5 & 26.5 & 92.4 & 66.4 & 40.8 & 54.0 & 58.42 \\
    \rowcolor{blue!6}\multirow{-3}{*}{Llama $\rightarrow$ Llama} & \textsc{XKV}   & \textbf{50.4} & \textbf{56.8} & \textbf{19.9} & \textbf{30.3} & \textbf{93.1} & \textbf{73.9} & \textbf{42.1} & \textbf{55.7} & \textbf{61.96} \\
    \bottomrule
  \end{tabular}
\vspace{-10pt}
\end{table*}
\endgroup

\paragraph{Parameter sharing and complexity.}
Direct pooling is linear in prompt length, $O(\sum_a L_aH_akN_aD_a)$, and the
cross-layer self-attention operates on only $L_R$ memory tokens. The position
decoder attends from each receiver position to those $L_R$ tokens rather than
performing self-attention over all $L_RN_R$ raw cache rows. Moreover, the
bottleneck, translation blocks, position decoder, gate, and output heads are
all shared across receiver layers, in contrast to \lcfx's independent
per-layer projectors, so the only depth-dependent parameters are the pooling
queries, the layer embeddings, and the layer map. Together these choices keep
the translator compact while retaining position-specific updates.

\begin{figure}[!t]
  \centering
  \includegraphics[width=\columnwidth]{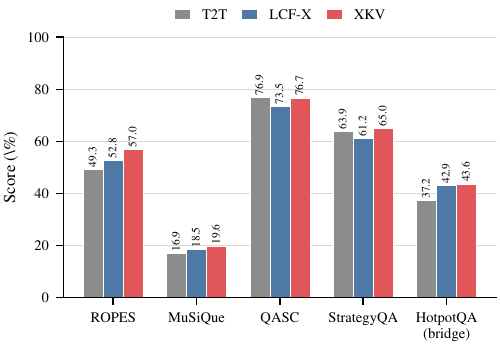}
  \vspace{-19pt}
  \caption{Dataset-level mean primary scores over the full $3\times3$
  sharer--receiver grid. Each dataset forms one group and each bar is a
  communication method; we use F1 for generative QA and accuracy for
  classification. XKV is best on four of the five datasets.}
  \label{fig:overall-comparison}
  \vspace{-17pt}
\end{figure}

\section{Experimental Setup}\label{sec:experiments}
\subsection{Task setting and models}
We evaluate communication in the \emph{cross-context} setting: the
\sharer{} and the \receiver{} each observe only part of the evidence, and the
receiver must combine its own context with information received from the
sharer. The question and, for classification tasks, the answer choices are
visible to both models; only the evidence is partitioned. We use a
deterministic per-example random assignment so that the two contexts are
disjoint without systematically assigning a particular evidence type to one
role. Specifically, ROPES assigns the background and situation passages one
per side; QASC assigns its two supporting facts one per side; 2-hop MuSiQue
and HotpotQA-bridge place one supporting paragraph on each side and divide
the distractors between them (four per side for HotpotQA); and StrategyQA
shuffles its supporting facts and splits them into two nonempty halves.
Thus, neither model receives all annotated supporting evidence.
We use three frozen decoder-only language models spanning three families,
Qwen3-0.6B \citep{yang2025qwen3}, Gemma-3-1B
\citep{gemmateam2025gemma3}, and Llama-3.2-3B
\citep{grattafiori2024llama3,meta2024llama32}, and evaluate the complete
$3\times3$ ordered grid in which any model may act as sharer or receiver,
yielding six heterogeneous off-diagonal pairings and three homogeneous
diagonal pairings. Collectively, the off-diagonal pairs span differences in
depth, attention geometry, and tokenizer; we assume neither token-level nor
layer-level correspondence. All reported numbers are means
over these nine pairings unless stated otherwise.

\subsection{Datasets}
We evaluate on five split-evidence benchmarks: \textbf{ROPES}
\citep{lin2019ropes}, \textbf{MuSiQue} \citep{trivedi2022musique}, and
\textbf{HotpotQA-bridge} \citep{yang2018hotpotqa} for generative multi-hop
question answering, and \textbf{QASC} \citep{khot2020qasc} and
\textbf{StrategyQA} \citep{geva2021strategyqa} for classification-style
reasoning. For each example, we partition the evidence so
that the sharer and receiver hold complementary contexts and the receiver
cannot solve the task reliably from its own context alone. For the generative
QA datasets we report exact match (EM) and F1; for the classification datasets
we report accuracy. In the main text, we use F1 for generative QA and accuracy
for classification as the primary score.

\subsection{Baselines}
We compare \xkv{} with two communication baselines.
\textbf{T2T} is a text-to-text pipeline: the sharer generates a
natural-language message from its private context and the question, and
the receiver answers from its own context plus that message. 
This is the
standard communication pattern in deployed agent systems
\citep{du2023debate,wu2023autogen,chen2024agentverse}, but it incurs
autoregressive message generation and re-encoding.
\textbf{LCF-X}, the cross-context extension of Latent Cache Flow
\citep{rossi2026latentcacheflow}, compresses the sharer cache into a
position-free summary injected into the receiver as gated KV residuals; it
is the most relevant latent baseline for our setting because it removes
the shared-token assumption of earlier cache-to-cache methods
\citep{fu2026cache,dery2026latent}. Since the original cross-context
design assumes matched layer count and KV dimensions and was evaluated
only on Qwen3-0.6B $\rightarrow$ Qwen3-0.6B, we strengthen it for the six
heterogeneous cells with our learned $L_S\!\rightarrow L_R$ layer-axis
map, a per-layer down-projection accepting concatenated sharer and
receiver features of different widths, and an up-projection emitting
receiver-native KV residuals; LCF-X in the experiments refers to this
project-authored heterogeneous extension.
\textbf{XKV} is our receiver-aware latent protocol, which builds a compact
joint memory from both caches and lets each receiver position retrieve its
own receiver-native residual update.


\subsection{Training and evaluation details}
For \xkv{} and \lcfx{}, only the lightweight communication module is
trained; all sharer and receiver weights remain frozen. Where applicable,
we use the same default recipe across datasets, with the optimizer,
learning-rate schedule, batch size, and decoding settings summarized in
the technical appendix. We measure two kinds of efficiency:
\emph{communication latency} covers only the learned latent module (the
fusor for \lcfx{}, the translator for \xkv{}), while \emph{end-to-end
latency} covers the full per-example pipeline of sharer forward pass,
communication step, and receiver-side answering. Since \ttt{} has neither
module, we report only its end-to-end latency, which includes
autoregressive message generation and receiver re-encoding. Every experiment is run with Nvidia A100 80GB, Ubuntu 22.04 LTS, and CUDA 12.2 once.

\section{Results}\label{sec:results}

\begingroup
\let\pairwisetextbf\textbf
\renewcommand{\textbf}[1]{\underline{\pairwisetextbf{#1}}}
\begin{table*}[t]
  \centering
  \small
  \renewcommand{\arraystretch}{.99}
  \setlength{\tabcolsep}{3.1pt}
  \caption{Latency results for same-model sharer--receiver settings.
  Each dataset reports latency in milliseconds per example; lower is better.
  Communication (Comm.) is fusor-only time for LCF-X and translator-only time
  for XKV. Since T2T has neither module, only its end-to-end (E2E) latency is
  reported. Cross-dataset Avg. averages the available measurement over five datasets.}
  \vspace{-5pt}
  \label{tab:full-latency-main}
  \begin{tabular}{llcccccccccccc}
    \toprule
    \multirow{2}{*}{Sharer $\rightarrow$ Receiver} & \multirow{2}{*}{Method}
      & \multicolumn{2}{c}{ROPES} & \multicolumn{2}{c}{MuSiQue}
      & \multicolumn{2}{c}{QASC} & \multicolumn{2}{c}{StrategyQA}
      & \multicolumn{2}{c}{HotpotQA-bridge}
      & \multicolumn{2}{c}{Cross-dataset Avg.} \\
    \cmidrule(lr){3-4}\cmidrule(lr){5-6}\cmidrule(lr){7-8}
    \cmidrule(lr){9-10}\cmidrule(lr){11-12}\cmidrule(lr){13-14}
      & & Comm. & E2E & Comm. & E2E & Comm. & E2E
      & Comm. & E2E & Comm. & E2E & Comm. & E2E \\
    \midrule
    \rowcolor{blue!6} & \textsc{T2T} & -- & 955 & -- & 974 & -- & 328 & -- & 848 & -- & 995 & -- & 820.0 \\
    \rowcolor{blue!6} & \textsc{LCF-X} & 31 & 165 & 115 & 376 & 42 & 156 & 42 & 154 & 64 & 259 & 58.8 & 222.0 \\
    \rowcolor{blue!6}\multirow{-3}{*}{Qwen $\rightarrow$ Qwen} & \textsc{XKV} & \textbf{6} & \textbf{139} & \textbf{6} & \textbf{218} & \textbf{6} & \textbf{119} & \textbf{10} & \textbf{122} & \textbf{6} & \textbf{201} & \textbf{6.8} & \textbf{159.8} \\
    \rowcolor{blue!6} & \textsc{T2T} & -- & 1549 & -- & 1225 & -- & 790 & -- & 1884 & -- & 1372 & -- & 1364.0 \\
    \rowcolor{blue!6} & \textsc{LCF-X} & 39 & 213 & 84 & 427 & 45 & 188 & 44 & 215 & 58 & 294 & 54.0 & 267.4 \\
    \rowcolor{blue!6}\multirow{-3}{*}{Gemma $\rightarrow$ Gemma} & \textsc{XKV} & \textbf{5} & \textbf{171} & \textbf{5} & \textbf{264} & \textbf{5} & \textbf{145} & \textbf{5} & \textbf{146} & \textbf{5} & \textbf{248} & \textbf{5.0} & \textbf{194.8} \\
    \rowcolor{blue!6} & \textsc{T2T} & -- & 1282 & -- & 1421 & -- & 842 & -- & 1296 & -- & 1377 & -- & 1243.6 \\
    \rowcolor{blue!6} & \textsc{LCF-X} & 41 & 147 & 123 & 325 & 46 & 135 & 46 & 139 & 64 & 215 & 64.0 & 192.2 \\
    \rowcolor{blue!6}\multirow{-3}{*}{Llama $\rightarrow$ Llama} & \textsc{XKV} & \textbf{6} & \textbf{111} & \textbf{6} & \textbf{223} & \textbf{6} & \textbf{94} & \textbf{6} & \textbf{105} & \textbf{6} & \textbf{174} & \textbf{6.0} & \textbf{141.4} \\
    \bottomrule
  \end{tabular}%
  \vspace{-5pt}
\end{table*}
\endgroup

\subsection{Overall comparison}
Table~\ref{tab:full-pairwise-main} reports the complete pair-level results for
all nine ordered sharer--receiver settings, and
Figure~\ref{fig:overall-comparison} aggregates them into dataset-level means
(F1 for generative QA, accuracy for classification). Per-dataset score
matrices and EM breakdowns are provided in the technical appendix.

\begin{figure}[!t]
  \centering
  \includegraphics[width=\columnwidth]{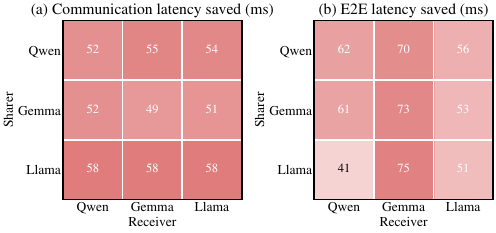}
  \vspace{-17pt}
  \caption{Pair-level communication and end-to-end latency savings relative to
  the faster baseline, averaged over five datasets. Rows denote sharers and
  columns receivers. (a) Communication latency saved by XKV (ms). (b)
  End-to-end latency saved by XKV. Positive values mean XKV is faster.}
  \label{fig:pairwise-heatmaps}
  \vspace{-10pt}
\end{figure}

\xkv{} is the strongest protocol overall: it attains the best or tied-best
cross-dataset average in eight of the nine settings and the best result in 31
of the 45 dataset--pair cells. It improves the \lcfx{} mean on every dataset
($+2.61$ macro points, up to $+4.2$ F1 on ROPES) and leads \ttt{} on four of
the five datasets ($+3.52$ macro points, including $+7.7$ F1 on ROPES and
$+6.4$ F1 on HotpotQA-bridge), trailing only on QASC by $0.27$ points.
Figure~\ref{fig:overall-comparison} also shows that the baselines are
complementary: \lcfx{} is stronger on the three generative QA datasets and
\ttt{} on the two classification datasets. \xkv{} is the only protocol
competitive in both regimes, which is why its advantage survives aggregation.
These quality gains cost nothing in latency: in every ordered pair,
Figure~\ref{fig:pairwise-heatmaps} shows \xkv{} saving $49$--$58$\,ms of
communication latency and $41$--$75$\,ms end to end relative to the faster
baseline, averaged over the five datasets.

\begin{table*}[t]
  \centering
  \small
  \setlength{\tabcolsep}{5pt}
  \caption{Ablation results over QASC, ROPES, and StrategyQA. Score changes are
  relative to full XKV and use accuracy for QASC and StrategyQA and F1 for
  ROPES. ``Comm.'' is translator-only latency.}
  \vspace{-8pt}
  \label{tab:ablation-main}
  \begin{tabular}{lrrrrrrr}
    \toprule
    Variant & QASC $\Delta$ & ROPES F1 $\Delta$ & StrategyQA $\Delta$ & Mean $\Delta$ & Comm. (ms) & E2E (ms) & Params \\
    \midrule
    Full XKV & 0.00 & 0.00 & 0.00 & 0.00 & 5.8 & 131.7 & 4.55M \\
    $-\mathrm{RP}$ & -0.41 & -0.82 & -0.16 & -0.46 & 4.7 & 128.7 & 3.58M \\
    $-\mathrm{RX}$ & -1.98 & -0.44 & -0.08 & -0.83 & 3.7 & 127.1 & 3.43M \\
    LCFP & -0.82 & -2.00 & +0.01 & -0.94 & 39.1 & 183.7 & 3.53M \\
    \bottomrule
  \end{tabular}
  \vspace{-10pt}
\end{table*}

\subsection{Efficiency}
Table~\ref{tab:full-latency-main} breaks down communication (Comm.) and
end-to-end (E2E) latency per dataset for the three same-model settings;
\ttt{} has no latent module, so only its E2E latency is defined. Latency
grids for all nine pairings appear in the technical appendix.

Two patterns stand out. First, the \xkv{} translator is nearly constant at
$5$--$10$\,ms in every cell, whereas the \lcfx{} fusor ranges from $31$ to
$123$\,ms and grows with context length, costing two to three times more on
MuSiQue than on QASC or StrategyQA. The reason is structural: direct pooling
is a single batched operation, while span pooling must process a number of
variable-length spans that scales with the prompt. Second, although both
latent methods pay the same two frozen prefills, \xkv{} is fastest end to end
in all fifteen cells, cutting E2E time by $26$--$28\%$ relative to \lcfx{};
\ttt{} is $5.1\times$ to $8.8\times$ slower than \xkv{} because
autoregressive message generation dominates its budget.
Aggregated over all 45 dataset--pair cells, the \xkv{} translator averages
$5.8$\,ms against $59.9$\,ms for the \lcfx{} fusor ($10.3\times$ faster),
lowers mean E2E latency from $227.9$ to $167.6$\,ms ($26.4\%$), runs
$6.8\times$ faster end to end than \ttt{}, and trains $76.1\%$ fewer
parameters than \lcfx{} ($4.55$M against $19.03$M). \xkv{} is thus
simultaneously the most accurate and the cheapest channel in our comparison.

\subsection{Ablation study}
We ablate \xkv{} on QASC, ROPES, and StrategyQA over all 27 matched
dataset--model-pair settings, removing one design decision at a time.
\textbf{$-\mathrm{RP}$} builds the memory from a sharer-only summary as in
\lcfx{}, testing whether \emph{what to communicate} should depend on the
receiver; \textbf{$-\mathrm{RX}$} replaces position-specific retrieval from
the raw receiver cache with broadcast layer memory, testing whether
receiver positions should explicitly query the joint memory;\textbf{LCFP} swaps our vectorized direct
pooling for \lcfx{}'s two-stage span pooling. Table~\ref{tab:ablation-main}
reports score changes relative to Full \xkv{} together with average latency
and parameter counts; per-pair matrices are in the technical appendix.
Both receiver-aware components earn their small cost. Removing receiver
pooling lowers the score on all three datasets ($-0.46$ on average) while
saving only $1.1$\,ms and $0.97$M parameters. Removing receiver
cross-attention costs more ($-0.83$ on average, $-1.98$ on QASC, where
evidence must be integrated selectively), showing that broadcast layer
memory cannot substitute for receiver-position queries into the joint
memory. 
Since neither removal
changes E2E latency by more than $3.5\%$, receiver-awareness is essentially
free at inference time.
LCFP shows that \xkv{}'s efficiency and quality are not in tension: it raises
translator latency from $5.8$ to $39.1$\,ms ($6.7\times$) and E2E latency by
$39.4\%$, yet also lowers the mean score by $0.94$ points, including $2.00$
F1 on ROPES. Direct pooling is thus not merely an implementation
optimization; replacing it makes the translator both slower and less
accurate.

\section{Conclusion}
We presented \xkv{}, a latent communication protocol for frozen,
heterogeneous language models holding complementary evidence. \xkv{}
decides \emph{what to communicate} jointly, pooling both KV caches into a
compact cross-layer memory, and lets every receiver cache position query
that memory for its own receiver-native residual. Across 45 settings (five
split-evidence datasets, six heterogeneous and three same-model ordered
pairs), \xkv{} attains the best overall score and improves on \lcfx{} on
every dataset, with a translator $10.3\times$ faster than the \lcfx{}
fusor, $76.1\%$ fewer trainable parameters, and end-to-end communication
$6.8\times$ faster than text exchange; ablations confirm that both
receiver-aware components add quality at negligible cost. Receiver-aware,
position-query cache communication thus removes the trade-off between
quality and efficiency. Future work includes scaling to larger models,
multi-turn and many-agent communication, and interpreting what the latent
messages transmit.

\bibliography{references}


+\begingroup
\let\pairwisetextbf\textbf
\renewcommand{\textbf}[1]{\underline{\pairwisetextbf{#1}}}
\begin{table*}[t]
  \centering
  \small
  \renewcommand{\arraystretch}{.99}
  \setlength{\tabcolsep}{2.1pt}
  \caption{Appendix backup: full latency results for all nine ordered sharer--receiver settings.
  Each dataset reports latency in milliseconds per example; lower is better.
  Communication (Comm.) is fusor-only time for LCF-X and translator-only time
  for XKV. Because T2T has neither module, only its end-to-end (E2E) latency is
  reported. Cross-dataset Avg. averages the available measurement over five datasets.}
  \label{tab:full-latency-appendix}
  \begin{tabular}{llcccccccccccc}
    \toprule
    \multirow{2}{*}{Sharer $\rightarrow$ Receiver} & \multirow{2}{*}{Method}
      & \multicolumn{2}{c}{ROPES} & \multicolumn{2}{c}{MuSiQue}
      & \multicolumn{2}{c}{QASC} & \multicolumn{2}{c}{StrategyQA}
      & \multicolumn{2}{c}{HotpotQA-bridge}
      & \multicolumn{2}{c}{Cross-dataset Avg.} \\
    \cmidrule(lr){3-4}\cmidrule(lr){5-6}\cmidrule(lr){7-8}
    \cmidrule(lr){9-10}\cmidrule(lr){11-12}\cmidrule(lr){13-14}
      & & Comm. & E2E & Comm. & E2E & Comm. & E2E
      & Comm. & E2E & Comm. & E2E & Comm. & E2E \\
    \midrule
    \rowcolor{blue!6} & \textsc{T2T} & -- & 955 & -- & 974 & -- & 328 & -- & 848 & -- & 995 & -- & 820.0 \\
    \rowcolor{blue!6} & \textsc{LCF-X} & 31 & 165 & 115 & 376 & 42 & 156 & 42 & 154 & 64 & 259 & 58.8 & 222.0 \\
    \rowcolor{blue!6}\multirow{-3}{*}{Qwen $\rightarrow$ Qwen} & \textsc{XKV} & \textbf{6} & \textbf{139} & \textbf{6} & \textbf{218} & \textbf{6} & \textbf{119} & \textbf{10} & \textbf{122} & \textbf{6} & \textbf{201} & \textbf{6.8} & \textbf{159.8} \\
    \multirow{3}{*}{Qwen $\rightarrow$ Gemma} & \textsc{T2T} & -- & 914 & -- & 971 & -- & 291 & -- & 855 & -- & 1001 & -- & 806.4 \\
     & \textsc{LCF-X} & 35 & 193 & 118 & 439 & 42 & 177 & 41 & 177 & 65 & 295 & 60.2 & 256.2 \\
     & \textsc{XKV} & \textbf{5} & \textbf{160} & \textbf{5} & \textbf{253} & \textbf{5} & \textbf{136} & \textbf{5} & \textbf{140} & \textbf{5} & \textbf{240} & \textbf{5.0} & \textbf{185.8} \\
    \rowcolor{blue!6} & \textsc{T2T} & -- & 927 & -- & 1006 & -- & 247 & -- & 807 & -- & 999 & -- & 797.2 \\
    \rowcolor{blue!6} & \textsc{LCF-X} & 36 & 148 & 116 & 331 & 42 & 139 & 42 & 139 & 65 & 224 & 60.2 & 196.2 \\
    \rowcolor{blue!6}\multirow{-3}{*}{Qwen $\rightarrow$ Llama} & \textsc{XKV} & \textbf{6} & \textbf{119} & \textbf{6} & \textbf{201} & \textbf{6} & \textbf{102} & \textbf{6} & \textbf{109} & \textbf{6} & \textbf{169} & \textbf{6.0} & \textbf{140.0} \\
    \multirow{3}{*}{Gemma $\rightarrow$ Qwen} & \textsc{T2T} & -- & 1605 & -- & 1236 & -- & 810 & -- & 1882 & -- & 1332 & -- & 1373.0 \\
     & \textsc{LCF-X} & 42 & 182 & 91 & 370 & 48 & 171 & 48 & 170 & 62 & 266 & 58.2 & 231.8 \\
     & \textsc{XKV} & \textbf{6} & \textbf{159} & \textbf{6} & \textbf{229} & \textbf{6} & \textbf{127} & \textbf{6} & \textbf{126} & \textbf{7} & \textbf{213} & \textbf{6.2} & \textbf{170.8} \\
    \rowcolor{blue!6} & \textsc{T2T} & -- & 1549 & -- & 1225 & -- & 790 & -- & 1884 & -- & 1372 & -- & 1364.0 \\
    \rowcolor{blue!6} & \textsc{LCF-X} & 39 & 213 & 84 & 427 & 45 & 188 & 44 & 215 & 58 & 294 & 54.0 & 267.4 \\
    \rowcolor{blue!6}\multirow{-3}{*}{Gemma $\rightarrow$ Gemma} & \textsc{XKV} & \textbf{5} & \textbf{171} & \textbf{5} & \textbf{264} & \textbf{5} & \textbf{145} & \textbf{5} & \textbf{146} & \textbf{5} & \textbf{248} & \textbf{5.0} & \textbf{194.8} \\
    \multirow{3}{*}{Gemma $\rightarrow$ Llama} & \textsc{T2T} & -- & 1566 & -- & 1274 & -- & 742 & -- & 1823 & -- & 1365 & -- & 1354.0 \\
     & \textsc{LCF-X} & 42 & 163 & 85 & 300 & 48 & 154 & 48 & 157 & 60 & 225 & 56.6 & 199.8 \\
     & \textsc{XKV} & \textbf{6} & \textbf{126} & \textbf{6} & \textbf{212} & \textbf{6} & \textbf{108} & \textbf{6} & \textbf{112} & \textbf{6} & \textbf{178} & \textbf{6.0} & \textbf{147.2} \\
    \rowcolor{blue!6} & \textsc{T2T} & -- & 1301 & -- & 1330 & -- & 925 & -- & 1344 & -- & 1387 & -- & 1257.4 \\
    \rowcolor{blue!6} & \textsc{LCF-X} & 42 & 171 & 120 & 372 & 47 & 155 & 48 & \textbf{157} & 64 & 257 & 64.2 & 222.4 \\
    \rowcolor{blue!6}\multirow{-3}{*}{Llama $\rightarrow$ Qwen} & \textsc{XKV} & \textbf{6} & \textbf{134} & \textbf{6} & \textbf{238} & \textbf{6} & \textbf{110} & \textbf{6} & 217 & \textbf{6} & \textbf{207} & \textbf{6.0} & \textbf{181.2} \\
    \multirow{3}{*}{Llama $\rightarrow$ Gemma} & \textsc{T2T} & -- & 1219 & -- & 1333 & -- & 880 & -- & 1361 & -- & 1407 & -- & 1240.0 \\
     & \textsc{LCF-X} & 40 & 197 & 118 & 435 & 46 & 176 & 47 & 213 & 64 & 293 & 63.0 & 262.8 \\
     & \textsc{XKV} & \textbf{5} & \textbf{153} & \textbf{5} & \textbf{278} & \textbf{5} & \textbf{133} & \textbf{5} & \textbf{134} & \textbf{5} & \textbf{239} & \textbf{5.0} & \textbf{187.4} \\
    \rowcolor{blue!6} & \textsc{T2T} & -- & 1282 & -- & 1421 & -- & 842 & -- & 1296 & -- & 1377 & -- & 1243.6 \\
    \rowcolor{blue!6} & \textsc{LCF-X} & 41 & 147 & 123 & 325 & 46 & 135 & 46 & 139 & 64 & 215 & 64.0 & 192.2 \\
    \rowcolor{blue!6}\multirow{-3}{*}{Llama $\rightarrow$ Llama} & \textsc{XKV} & \textbf{6} & \textbf{111} & \textbf{6} & \textbf{223} & \textbf{6} & \textbf{94} & \textbf{6} & \textbf{105} & \textbf{6} & \textbf{174} & \textbf{6.0} & \textbf{141.4} \\
    \bottomrule
  \end{tabular}%
\end{table*}
\endgroup

\begin{table*}[h]
  \centering
  \caption{Default training / inference hyperparameters.}
  \label{tab:hparams}
  \begin{tabular}{ll}
    \toprule
    Setting & Value \\
    \midrule
    Optimizer & AdamW \\
    Learning rate & $1\times10^{-4}$ \\
    LR schedule & linear warmup ($10\%$ of total steps) \\
    Effective batch size & 256 \\
    Gradient clipping & $1.0$ \\
    Pooling spans $P$ & $4$ \\
    Translator bottleneck dim & $128$ \\
    Validation holdout & $2.8\%$ of train (when no explicit eval split) \\
    Receiver max new tokens & $64$ \\
    Seed & $0$ \\
    \bottomrule
  \end{tabular}
\end{table*}



\begin{table*}[t]
  \centering
  \caption{ROPES EM/F1 scores (\%). Higher is better; cell-wise best values are bold. Mean across the nine model pairs: \textsc{LCF-X} 45.86/52.79; \textsc{XKV} 50.48/56.99; \textsc{T2T} 40.74/49.31.}
  \label{tab:methods-ropes-scores}
  \begin{tabular}{llccc}
    \toprule
    Sharer & Receiver & \textsc{LCF-X} & \textsc{XKV} & \textsc{T2T} \\
    \midrule
    Qwen-0.6B & Qwen-0.6B & 45.2/52.4 & \textbf{47.9}/\textbf{55.6} & 35.8/45.2 \\
    Qwen-0.6B & Gemma-1B & 43.9/51.3 & \textbf{46.2}/\textbf{54.3} & 42.4/48.6 \\
    Qwen-0.6B & Llama-3B & 51.4/58.5 & \textbf{57.3}/\textbf{61.8} & 50.6/58.0 \\
    Gemma-1B & Qwen-0.6B & 47.2/53.7 & \textbf{49.1}/\textbf{55.4} & 31.9/42.5 \\
    Gemma-1B & Gemma-1B & 45.5/51.4 & \textbf{48.1}/\textbf{55.6} & 31.8/42.1 \\
    Gemma-1B & Llama-3B & 42.1/49.8 & \textbf{54.0}/\textbf{58.7} & 46.9/55.7 \\
    Llama-3B & Qwen-0.6B & 46.7/54.8 & \textbf{53.0}/\textbf{59.0} & 38.6/47.7 \\
    Llama-3B & Gemma-1B & 45.1/50.4 & \textbf{48.3}/\textbf{55.7} & 40.3/48.3 \\
    Llama-3B & Llama-3B & 45.6/52.8 & \textbf{50.4}/\textbf{56.8} & 48.4/55.7 \\
    \bottomrule
  \end{tabular}
\end{table*}

\begin{table*}[t]
  \centering
  \caption{ROPES latency in milliseconds per example. Communication is measured only for the latent modules (LCF-X fusor and XKV translator); T2T therefore reports end-to-end latency only. Lower is better; cell-wise minima at the displayed precision are bold. Means across the nine model pairs: \textsc{LCF-X} 38.7/175.4; \textsc{XKV} 5.7/141.3; \textsc{T2T} E2E 1257.6.}
  \label{tab:methods-ropes-latency}
  \begin{tabular}{llccccc}
    \toprule
    \multirow{2}{*}{Sharer} & \multirow{2}{*}{Receiver} & \multicolumn{2}{c}{\textsc{LCF-X}} & \multicolumn{2}{c}{\textsc{XKV}} & \textsc{T2T} \\
    \cmidrule(lr){3-4}\cmidrule(lr){5-6}
      & & Comm. & E2E & Comm. & E2E & E2E \\
    \midrule
    Qwen-0.6B & Qwen-0.6B & 31 & 165 & \textbf{6} & \textbf{139} & 955 \\
    Qwen-0.6B & Gemma-1B & 35 & 193 & \textbf{5} & \textbf{160} & 914 \\
    Qwen-0.6B & Llama-3B & 36 & 148 & \textbf{6} & \textbf{119} & 927 \\
    Gemma-1B & Qwen-0.6B & 42 & 182 & \textbf{6} & \textbf{159} & 1605 \\
    Gemma-1B & Gemma-1B & 39 & 213 & \textbf{5} & \textbf{171} & 1549 \\
    Gemma-1B & Llama-3B & 42 & 163 & \textbf{6} & \textbf{126} & 1566 \\
    Llama-3B & Qwen-0.6B & 42 & 171 & \textbf{6} & \textbf{134} & 1301 \\
    Llama-3B & Gemma-1B & 40 & 197 & \textbf{5} & \textbf{153} & 1219 \\
    Llama-3B & Llama-3B & 41 & 147 & \textbf{6} & \textbf{111} & 1282 \\
    \bottomrule
  \end{tabular}
\end{table*}

\begin{table*}[t]
  \centering
  \caption{MuSiQue EM/F1 scores (\%). Higher is better; cell-wise best values are bold. Mean across the nine model pairs: \textsc{LCF-X} 10.00/18.47; \textsc{XKV} 11.07/19.59; \textsc{T2T} 7.99/16.92.}
  \label{tab:methods-musique-scores}
  \begin{tabular}{llccc}
    \toprule
    Sharer & Receiver & \textsc{LCF-X} & \textsc{XKV} & \textsc{T2T} \\
    \midrule
    Qwen-0.6B & Qwen-0.6B & 7.7/15.4 & \textbf{8.1}/\textbf{16.0} & 4.6/12.2 \\
    Qwen-0.6B & Gemma-1B & 6.0/13.3 & 5.4/12.5 & \textbf{6.8}/\textbf{14.5} \\
    Qwen-0.6B & Llama-3B & 16.0/26.3 & \textbf{19.2}/\textbf{29.9} & 9.3/19.5 \\
    Gemma-1B & Qwen-0.6B & \textbf{8.4}/\textbf{16.6} & 8.1/15.8 & 3.4/10.8 \\
    Gemma-1B & Gemma-1B & 4.6/11.9 & 5.8/13.1 & \textbf{6.2}/\textbf{13.4} \\
    Gemma-1B & Llama-3B & 17.8/28.3 & \textbf{19.4}/\textbf{30.0} & 9.4/18.3 \\
    Llama-3B & Qwen-0.6B & 7.6/15.5 & \textbf{8.5}/\textbf{16.5} & 5.9/15.4 \\
    Llama-3B & Gemma-1B & 5.4/12.4 & 5.2/12.2 & \textbf{12.1}/\textbf{21.9} \\
    Llama-3B & Llama-3B & 16.5/26.5 & \textbf{19.9}/\textbf{30.3} & 14.2/26.3 \\
    \bottomrule
  \end{tabular}
\end{table*}

\begin{table*}[t]
  \centering
  \caption{MuSiQue latency in milliseconds per example. Communication is measured only for the latent modules (LCF-X fusor and XKV translator); T2T therefore reports end-to-end latency only. Lower is better; cell-wise minima at the displayed precision are bold. Means across the nine model pairs: \textsc{LCF-X} 107.8/375.0; \textsc{XKV} 5.7/235.1; \textsc{T2T} E2E 1196.7.}
  \label{tab:methods-musique-latency}
  \begin{tabular}{llccccc}
    \toprule
    \multirow{2}{*}{Sharer} & \multirow{2}{*}{Receiver} & \multicolumn{2}{c}{\textsc{LCF-X}} & \multicolumn{2}{c}{\textsc{XKV}} & \textsc{T2T} \\
    \cmidrule(lr){3-4}\cmidrule(lr){5-6}
      & & Comm. & E2E & Comm. & E2E & E2E \\
    \midrule
    Qwen-0.6B & Qwen-0.6B & 115 & 376 & \textbf{6} & \textbf{218} & 974 \\
    Qwen-0.6B & Gemma-1B & 118 & 439 & \textbf{5} & \textbf{253} & 971 \\
    Qwen-0.6B & Llama-3B & 116 & 331 & \textbf{6} & \textbf{201} & 1006 \\
    Gemma-1B & Qwen-0.6B & 91 & 370 & \textbf{6} & \textbf{229} & 1236 \\
    Gemma-1B & Gemma-1B & 84 & 427 & \textbf{5} & \textbf{264} & 1225 \\
    Gemma-1B & Llama-3B & 85 & 300 & \textbf{6} & \textbf{212} & 1274 \\
    Llama-3B & Qwen-0.6B & 120 & 372 & \textbf{6} & \textbf{238} & 1330 \\
    Llama-3B & Gemma-1B & 118 & 435 & \textbf{5} & \textbf{278} & 1333 \\
    Llama-3B & Llama-3B & 123 & 325 & \textbf{6} & \textbf{223} & 1421 \\
    \bottomrule
  \end{tabular}
\end{table*}

\begin{table*}[t]
  \centering
  \caption{QASC accuracy scores (\%). Higher is better; cell-wise best values are bold. Mean across the nine model pairs: \textsc{LCF-X} 73.44; \textsc{XKV} 76.67; \textsc{T2T} 76.93.}
  \label{tab:methods-qasc-scores}
  \begin{tabular}{llccc}
    \toprule
    Sharer & Receiver & \textsc{LCF-X} & \textsc{XKV} & \textsc{T2T} \\
    \midrule
    Qwen-0.6B & Qwen-0.6B & \textbf{82.4} & 82.2 & 78.6 \\
    Qwen-0.6B & Gemma-1B & 49.2 & 55.1 & \textbf{65.8} \\
    Qwen-0.6B & Llama-3B & 92.0 & \textbf{93.3} & 86.7 \\
    Gemma-1B & Qwen-0.6B & 81.3 & 82.4 & \textbf{84.6} \\
    Gemma-1B & Gemma-1B & 44.8 & 53.9 & \textbf{55.7} \\
    Gemma-1B & Llama-3B & 92.9 & \textbf{93.0} & 89.4 \\
    Llama-3B & Qwen-0.6B & 81.1 & 82.1 & \textbf{85.0} \\
    Llama-3B & Gemma-1B & 44.9 & 54.9 & \textbf{55.9} \\
    Llama-3B & Llama-3B & 92.4 & \textbf{93.1} & 90.7 \\
    \bottomrule
  \end{tabular}
\end{table*}

\begin{table*}[t]
  \centering
  \caption{QASC latency in milliseconds per example. Communication is measured only for the latent modules (LCF-X fusor and XKV translator); T2T therefore reports end-to-end latency only. Lower is better; cell-wise minima at the displayed precision are bold. Means across the nine model pairs: \textsc{LCF-X} 45.1/161.2; \textsc{XKV} 5.7/119.3; \textsc{T2T} E2E 650.6.}
  \label{tab:methods-qasc-latency}
  \begin{tabular}{llccccc}
    \toprule
    \multirow{2}{*}{Sharer} & \multirow{2}{*}{Receiver} & \multicolumn{2}{c}{\textsc{LCF-X}} & \multicolumn{2}{c}{\textsc{XKV}} & \textsc{T2T} \\
    \cmidrule(lr){3-4}\cmidrule(lr){5-6}
      & & Comm. & E2E & Comm. & E2E & E2E \\
    \midrule
    Qwen-0.6B & Qwen-0.6B & 42 & 156 & \textbf{6} & \textbf{119} & 328 \\
    Qwen-0.6B & Gemma-1B & 42 & 177 & \textbf{5} & \textbf{136} & 291 \\
    Qwen-0.6B & Llama-3B & 42 & 139 & \textbf{6} & \textbf{102} & 247 \\
    Gemma-1B & Qwen-0.6B & 48 & 171 & \textbf{6} & \textbf{127} & 810 \\
    Gemma-1B & Gemma-1B & 45 & 188 & \textbf{5} & \textbf{145} & 790 \\
    Gemma-1B & Llama-3B & 48 & 154 & \textbf{6} & \textbf{108} & 742 \\
    Llama-3B & Qwen-0.6B & 47 & 155 & \textbf{6} & \textbf{110} & 925 \\
    Llama-3B & Gemma-1B & 46 & 176 & \textbf{5} & \textbf{133} & 880 \\
    Llama-3B & Llama-3B & 46 & 135 & \textbf{6} & \textbf{94} & 842 \\
    \bottomrule
  \end{tabular}
\end{table*}

\begin{table*}[t]
  \centering
  \caption{StrategyQA accuracy scores (\%). Higher is better; cell-wise best values are bold. Mean across the nine model pairs: \textsc{LCF-X} 61.17; \textsc{XKV} 65.01; \textsc{T2T} 63.91.}
  \label{tab:methods-strategyqa-scores}
  \begin{tabular}{llccc}
    \toprule
    Sharer & Receiver & \textsc{LCF-X} & \textsc{XKV} & \textsc{T2T} \\
    \midrule
    Qwen-0.6B & Qwen-0.6B & 58.8 & 61.4 & \textbf{64.1} \\
    Qwen-0.6B & Gemma-1B & \textbf{60.8} & 58.2 & 51.9 \\
    Qwen-0.6B & Llama-3B & 72.9 & \textbf{73.6} & 73.4 \\
    Gemma-1B & Qwen-0.6B & 59.3 & 62.0 & \textbf{65.9} \\
    Gemma-1B & Gemma-1B & 58.2 & \textbf{58.8} & 57.6 \\
    Gemma-1B & Llama-3B & 68.8 & \textbf{74.1} & 70.9 \\
    Llama-3B & Qwen-0.6B & 53.1 & \textbf{65.8} & 65.6 \\
    Llama-3B & Gemma-1B & 52.2 & \textbf{57.3} & 53.7 \\
    Llama-3B & Llama-3B & 66.4 & \textbf{73.9} & 72.1 \\
    \bottomrule
  \end{tabular}
\end{table*}

\begin{table*}[t]
  \centering
  \caption{StrategyQA latency in milliseconds per example. Communication is measured only for the latent modules (LCF-X fusor and XKV translator); T2T therefore reports end-to-end latency only. Lower is better; cell-wise minima at the displayed precision are bold. Means across the nine model pairs: \textsc{LCF-X} 45.1/169.0; \textsc{XKV} 6.1/134.6; \textsc{T2T} E2E 1344.4.}
  \label{tab:methods-strategyqa-latency}
  \begin{tabular}{llccccc}
    \toprule
    \multirow{2}{*}{Sharer} & \multirow{2}{*}{Receiver} & \multicolumn{2}{c}{\textsc{LCF-X}} & \multicolumn{2}{c}{\textsc{XKV}} & \textsc{T2T} \\
    \cmidrule(lr){3-4}\cmidrule(lr){5-6}
      & & Comm. & E2E & Comm. & E2E & E2E \\
    \midrule
    Qwen-0.6B & Qwen-0.6B & 42 & 154 & \textbf{10} & \textbf{122} & 848 \\
    Qwen-0.6B & Gemma-1B & 41 & 177 & \textbf{5} & \textbf{140} & 855 \\
    Qwen-0.6B & Llama-3B & 42 & 139 & \textbf{6} & \textbf{109} & 807 \\
    Gemma-1B & Qwen-0.6B & 48 & 170 & \textbf{6} & \textbf{126} & 1882 \\
    Gemma-1B & Gemma-1B & 44 & 215 & \textbf{5} & \textbf{146} & 1884 \\
    Gemma-1B & Llama-3B & 48 & 157 & \textbf{6} & \textbf{112} & 1823 \\
    Llama-3B & Qwen-0.6B & 48 & \textbf{157} & \textbf{6} & 217 & 1344 \\
    Llama-3B & Gemma-1B & 47 & 213 & \textbf{5} & \textbf{134} & 1361 \\
    Llama-3B & Llama-3B & 46 & 139 & \textbf{6} & \textbf{105} & 1296 \\
    \bottomrule
  \end{tabular}
\end{table*}

\begin{table*}[t]
  \centering
  \caption{HotpotQA-bridge EM/F1 scores (\%). Higher is better; cell-wise best values are bold. Mean across the nine model pairs: \textsc{LCF-X} 31.06/42.96; \textsc{XKV} 31.72/43.61; \textsc{T2T} 24.72/37.20.}
  \label{tab:methods-hotpotqa-bridge-scores}
  \begin{tabular}{llccc}
    \toprule
    Sharer & Receiver & \textsc{LCF-X} & \textsc{XKV} & \textsc{T2T} \\
    \midrule
    Qwen-0.6B & Qwen-0.6B & 29.1/40.6 & \textbf{29.5}/\textbf{40.7} & 16.6/27.1 \\
    Qwen-0.6B & Gemma-1B & 23.3/34.0 & \textbf{23.8}/\textbf{34.7} & 22.4/34.0 \\
    Qwen-0.6B & Llama-3B & 41.1/54.5 & \textbf{41.5}/\textbf{55.0} & 30.4/44.0 \\
    Gemma-1B & Qwen-0.6B & 29.4/40.9 & \textbf{29.5}/\textbf{41.0} & 15.5/25.7 \\
    Gemma-1B & Gemma-1B & 23.5/\textbf{34.4} & \textbf{23.9}/\textbf{34.4} & 21.7/33.1 \\
    Gemma-1B & Llama-3B & 40.2/53.5 & \textbf{41.5}/\textbf{55.1} & 30.4/44.5 \\
    Llama-3B & Qwen-0.6B & 28.6/40.3 & \textbf{29.3}/\textbf{40.9} & 21.1/32.9 \\
    Llama-3B & Gemma-1B & 23.5/34.4 & 24.4/35.0 & \textbf{28.3}/\textbf{41.5} \\
    Llama-3B & Llama-3B & 40.8/54.0 & \textbf{42.1}/\textbf{55.7} & 36.1/52.0 \\
    \bottomrule
  \end{tabular}
\end{table*}

\begin{table*}[t]
  \centering
  \caption{HotpotQA-bridge latency in milliseconds per example. Communication is measured only for the latent modules (LCF-X fusor and XKV translator); T2T therefore reports end-to-end latency only. Lower is better; cell-wise minima at the displayed precision are bold. Means across the nine model pairs: \textsc{LCF-X} 62.9/258.7; \textsc{XKV} 5.8/207.7; \textsc{T2T} E2E 1248.3.}
  \label{tab:methods-hotpotqa-bridge-latency}
  \begin{tabular}{llccccc}
    \toprule
    \multirow{2}{*}{Sharer} & \multirow{2}{*}{Receiver} & \multicolumn{2}{c}{\textsc{LCF-X}} & \multicolumn{2}{c}{\textsc{XKV}} & \textsc{T2T} \\
    \cmidrule(lr){3-4}\cmidrule(lr){5-6}
      & & Comm. & E2E & Comm. & E2E & E2E \\
    \midrule
    Qwen-0.6B & Qwen-0.6B & 64 & 259 & \textbf{6} & \textbf{201} & 995 \\
    Qwen-0.6B & Gemma-1B & 65 & 295 & \textbf{5} & \textbf{240} & 1001 \\
    Qwen-0.6B & Llama-3B & 65 & 224 & \textbf{6} & \textbf{169} & 999 \\
    Gemma-1B & Qwen-0.6B & 62 & 266 & \textbf{7} & \textbf{213} & 1332 \\
    Gemma-1B & Gemma-1B & 58 & 294 & \textbf{5} & \textbf{248} & 1372 \\
    Gemma-1B & Llama-3B & 60 & 225 & \textbf{6} & \textbf{178} & 1365 \\
    Llama-3B & Qwen-0.6B & 64 & 257 & \textbf{6} & \textbf{207} & 1387 \\
    Llama-3B & Gemma-1B & 64 & 293 & \textbf{5} & \textbf{239} & 1407 \\
    Llama-3B & Llama-3B & 64 & 215 & \textbf{6} & \textbf{174} & 1377 \\
    \bottomrule
  \end{tabular}
\end{table*}


\begin{table*}[t]
  \centering
  \caption{QASC accuracy scores (\%). Higher is better; cell-wise best values are bold. Mean across the nine model pairs: Full 76.67; $-\mathrm{RP}$ 76.26; $-\mathrm{RX}$ 74.69; LCFP 75.84.}
  \label{tab:ablation-qasc-scores}
  \begin{tabular}{llcccc}
    \toprule
    Sharer & Receiver & Full & $-\mathrm{RP}$ & $-\mathrm{RX}$ & LCFP \\
    \midrule
    Qwen-0.6B & Qwen-0.6B & \textbf{82.2} & 81.6 & 80.8 & 81.7 \\
    Qwen-0.6B & Gemma-1B & \textbf{55.1} & 53.8 & 49.6 & 51.9 \\
    Qwen-0.6B & Llama-3B & \textbf{93.3} & 93.1 & 92.8 & 93.1 \\
    Gemma-1B & Qwen-0.6B & \textbf{82.4} & 81.9 & 81.3 & 81.3 \\
    Gemma-1B & Gemma-1B & 53.9 & 52.3 & 49.6 & \textbf{54.0} \\
    Gemma-1B & Llama-3B & 93.0 & \textbf{93.6} & 93.2 & 93.5 \\
    Llama-3B & Qwen-0.6B & \textbf{82.1} & 81.9 & 81.1 & 80.9 \\
    Llama-3B & Gemma-1B & \textbf{54.9} & 54.6 & 51.0 & 53.2 \\
    Llama-3B & Llama-3B & 93.1 & \textbf{93.5} & 92.8 & 93.0 \\
    \bottomrule
  \end{tabular}
\end{table*}

\begin{table*}[t]
  \centering
  \caption{QASC ablation latency in milliseconds per example. Entries are translator/total; lower is better and cell-wise minima are bold. Mean across the nine model pairs: Full 5.7/119.3; $-\mathrm{RP}$ 4.7/120.2; $-\mathrm{RX}$ 3.7/118.4; LCFP 37.4/171.3.}
  \label{tab:ablation-qasc-latency}
  \begin{tabular}{llcccc}
    \toprule
    Sharer & Receiver & Full & $-\mathrm{RP}$ & $-\mathrm{RX}$ & LCFP \\
    \midrule
    Qwen-0.6B & Qwen-0.6B & 6/119 & 5/118 & \textbf{4}/\textbf{115} & 48/320 \\
    Qwen-0.6B & Gemma-1B & 5/\textbf{136} & 4/139 & \textbf{3}/138 & 31/167 \\
    Qwen-0.6B & Llama-3B & 6/102 & 5/102 & \textbf{4}/\textbf{101} & 33/130 \\
    Gemma-1B & Qwen-0.6B & 6/127 & 5/128 & \textbf{4}/\textbf{123} & 39/161 \\
    Gemma-1B & Gemma-1B & 5/\textbf{145} & 4/146 & \textbf{3}/147 & 36/178 \\
    Gemma-1B & Llama-3B & 6/108 & 5/110 & \textbf{4}/\textbf{106} & 38/146 \\
    Llama-3B & Qwen-0.6B & 6/110 & 5/110 & \textbf{4}/\textbf{109} & 38/146 \\
    Llama-3B & Gemma-1B & 5/133 & 4/134 & \textbf{3}/\textbf{132} & 36/166 \\
    Llama-3B & Llama-3B & 6/\textbf{94} & 5/95 & \textbf{4}/95 & 38/128 \\
    \bottomrule
  \end{tabular}
\end{table*}

\begin{table*}[t]
  \centering
  \caption{ROPES EM/F1 scores (\%). Higher is better; cell-wise best values are bold. Mean across the nine model pairs: Full 50.48/56.99; $-\mathrm{RP}$ 49.38/56.17; $-\mathrm{RX}$ 49.79/56.54; LCFP 47.77/54.99.}
  \label{tab:ablation-ropes-scores}
  \begin{tabular}{llcccc}
    \toprule
    Sharer & Receiver & Full & $-\mathrm{RP}$ & $-\mathrm{RX}$ & LCFP \\
    \midrule
    Qwen-0.6B & Qwen-0.6B & 47.9/55.6 & \textbf{49.5}/\textbf{56.6} & 46.3/53.5 & 45.6/52.7 \\
    Qwen-0.6B & Gemma-1B & 46.2/\textbf{54.3} & \textbf{47.0}/54.1 & 45.5/53.7 & 46.6/53.9 \\
    Qwen-0.6B & Llama-3B & \textbf{57.3}/\textbf{61.8} & 53.3/59.2 & 55.6/61.2 & 53.7/60.3 \\
    Gemma-1B & Qwen-0.6B & \textbf{49.1}/\textbf{55.4} & 45.6/52.1 & 44.9/51.7 & 44.7/53.6 \\
    Gemma-1B & Gemma-1B & \textbf{48.1}/\textbf{55.6} & 46.7/54.2 & 47.0/53.8 & 44.0/51.2 \\
    Gemma-1B & Llama-3B & 54.0/58.7 & 52.8/58.1 & \textbf{57.5}/\textbf{62.9} & 52.0/58.3 \\
    Llama-3B & Qwen-0.6B & \textbf{53.0}/\textbf{59.0} & 48.9/57.3 & 45.7/53.9 & 45.3/53.2 \\
    Llama-3B & Gemma-1B & 48.3/\textbf{55.7} & \textbf{48.5}/55.6 & 47.0/54.8 & 46.3/53.5 \\
    Llama-3B & Llama-3B & 50.4/56.8 & 52.1/58.3 & \textbf{58.6}/\textbf{63.4} & 51.7/58.2 \\
    \bottomrule
  \end{tabular}
\end{table*}

\begin{table*}[t]
  \centering
  \caption{ROPES ablation latency in milliseconds per example. Entries are translator/total; lower is better and cell-wise minima are bold. Mean across the nine model pairs: Full 5.7/141.3; $-\mathrm{RP}$ 4.8/144.6; $-\mathrm{RX}$ 3.7/141.1; LCFP 37.4/177.9.}
  \label{tab:ablation-ropes-latency}
  \begin{tabular}{llcccc}
    \toprule
    Sharer & Receiver & Full & $-\mathrm{RP}$ & $-\mathrm{RX}$ & LCFP \\
    \midrule
    Qwen-0.6B & Qwen-0.6B & 6/\textbf{139} & 5/\textbf{139} & \textbf{4}/140 & 34/171 \\
    Qwen-0.6B & Gemma-1B & 5/\textbf{160} & 4/163 & \textbf{3}/167 & 33/194 \\
    Qwen-0.6B & Llama-3B & 6/119 & 5/118 & \textbf{4}/\textbf{116} & 34/147 \\
    Gemma-1B & Qwen-0.6B & 6/159 & 6/153 & \textbf{4}/\textbf{147} & 39/186 \\
    Gemma-1B & Gemma-1B & 5/171 & 4/\textbf{169} & \textbf{3}/\textbf{169} & 38/210 \\
    Gemma-1B & Llama-3B & 6/\textbf{126} & 5/127 & \textbf{4}/127 & 40/167 \\
    Llama-3B & Qwen-0.6B & 6/\textbf{134} & 5/138 & \textbf{4}/\textbf{134} & 40/175 \\
    Llama-3B & Gemma-1B & 5/\textbf{153} & 4/174 & \textbf{3}/158 & 39/200 \\
    Llama-3B & Llama-3B & 6/\textbf{111} & 5/120 & \textbf{4}/112 & 40/151 \\
    \bottomrule
  \end{tabular}
\end{table*}

\begin{table*}[t]
  \centering
  \caption{StrategyQA accuracy scores (\%). Higher is better; cell-wise best values are bold. Mean across the nine model pairs: Full 65.01; $-\mathrm{RP}$ 64.86; $-\mathrm{RX}$ 64.93; LCFP 65.02.}
  \label{tab:ablation-strategyqa-scores}
  \begin{tabular}{llcccc}
    \toprule
    Sharer & Receiver & Full & $-\mathrm{RP}$ & $-\mathrm{RX}$ & LCFP \\
    \midrule
    Qwen-0.6B & Qwen-0.6B & 61.4 & \textbf{63.2} & 61.7 & \textbf{63.2} \\
    Qwen-0.6B & Gemma-1B & 58.2 & 58.2 & 57.3 & \textbf{59.1} \\
    Qwen-0.6B & Llama-3B & 73.6 & 72.1 & \textbf{73.8} & 72.6 \\
    Gemma-1B & Qwen-0.6B & 62.0 & 63.2 & 61.7 & \textbf{64.4} \\
    Gemma-1B & Gemma-1B & 58.8 & \textbf{59.9} & 57.9 & 57.9 \\
    Gemma-1B & Llama-3B & 74.1 & 73.0 & \textbf{77.1} & 73.6 \\
    Llama-3B & Qwen-0.6B & \textbf{65.8} & 62.6 & 61.7 & 62.6 \\
    Llama-3B & Gemma-1B & 57.3 & 58.5 & 58.5 & \textbf{59.1} \\
    Llama-3B & Llama-3B & 73.9 & 73.0 & \textbf{74.7} & 72.7 \\
    \bottomrule
  \end{tabular}
\end{table*}

\begin{table*}[t]
  \centering
  \caption{StrategyQA ablation latency in milliseconds per example. Entries are translator/total; lower is better and cell-wise minima are bold. Mean across the nine model pairs: Full 6.1/134.6; $-\mathrm{RP}$ 4.8/121.2; $-\mathrm{RX}$ 3.7/121.7; LCFP 42.4/201.9.}
  \label{tab:ablation-strategyqa-latency}
  \begin{tabular}{llcccc}
    \toprule
    Sharer & Receiver & Full & $-\mathrm{RP}$ & $-\mathrm{RX}$ & LCFP \\
    \midrule
    Qwen-0.6B & Qwen-0.6B & 10/122 & 5/\textbf{118} & \textbf{4}/122 & 78/564 \\
    Qwen-0.6B & Gemma-1B & 5/140 & 4/140 & \textbf{3}/\textbf{139} & 32/169 \\
    Qwen-0.6B & Llama-3B & 6/109 & 5/\textbf{103} & \textbf{4}/104 & 34/134 \\
    Gemma-1B & Qwen-0.6B & 6/126 & 5/\textbf{123} & \textbf{4}/128 & 38/161 \\
    Gemma-1B & Gemma-1B & 5/\textbf{146} & 4/148 & \textbf{3}/149 & 36/179 \\
    Gemma-1B & Llama-3B & 6/112 & 6/116 & \textbf{4}/\textbf{108} & 43/152 \\
    Llama-3B & Qwen-0.6B & 6/217 & 5/\textbf{111} & \textbf{4}/114 & 42/150 \\
    Llama-3B & Gemma-1B & 5/\textbf{134} & 4/135 & \textbf{3}/136 & 37/167 \\
    Llama-3B & Llama-3B & 6/105 & 5/97 & \textbf{4}/\textbf{95} & 42/141 \\
    \bottomrule
  \end{tabular}
\end{table*}


\begin{table*}[t]
  \centering
  \caption{Overall score comparison on five datasets (45 model-pair cells). The native score is F1 for generative QA and accuracy for classification. Tied best values at the displayed precision are counted for each tied method.}
  \label{tab:method-score-summary}
  \begin{tabular}{lcccc}
    \toprule
    Method & Macro native score & Best datasets & Best/tied-best cells & Avg. rank \\
    \midrule
    \textsc{LCF-X} & 49.76 & 0/5 & 4/45 & 2.29 \\
    \textsc{XKV} & 52.37 & 4/5 & 31/45 & 1.37 \\
    \textsc{T2T} & 48.86 & 1/5 & 11/45 & 2.34 \\
    \bottomrule
  \end{tabular}
\end{table*}

\begin{table*}[t]
  \centering
  \caption{Overall efficiency on five datasets (45 model-pair cells). Communication is fusor-only latency for LCF-X and translator-only latency for XKV; it is not defined for T2T. XKV is 26.4\% faster end-to-end than LCF-X, 6.8$\times$ faster than T2T, and uses 76.1\% fewer trainable parameters than LCF-X.}
  \label{tab:method-efficiency-summary}
  \begin{tabular}{lcccc}
    \toprule
    Method & Communication (ms) & E2E (ms) & E2E relative to XKV & Trainable params \\
    \midrule
    \textsc{LCF-X} & 59.9 & 227.9 & 1.36$\times$ & 19.03M \\
    \textsc{XKV} & 5.8 & 167.6 & 1.00$\times$ & 4.55M \\
    \textsc{T2T} & -- & 1139.5 & 6.80$\times$ & 0 \\
    \bottomrule
  \end{tabular}
\end{table*}

\begin{table*}[t]
  \centering
  \caption{Per-dataset XKV improvements recomputed from the displayed detailed entries. Score deltas use F1 for generative QA and accuracy for classification; positive values favor XKV. Speedups are ratios of baseline E2E latency to XKV E2E latency.}
  \label{tab:xkv-improvements}
  \begin{tabular}{lrrrr}
    \toprule
    Dataset & $\Delta$ vs. LCF-X & $\Delta$ vs. T2T & Speedup vs. LCF-X & Speedup vs. T2T \\
    \midrule
    ROPES & +4.20 & +7.68 & 1.24$\times$ & 8.90$\times$ \\
    MuSiQue & +1.12 & +2.67 & 1.59$\times$ & 5.09$\times$ \\
    QASC & +3.22 & -0.27 & 1.35$\times$ & 5.45$\times$ \\
    StrategyQA & +3.84 & +1.10 & 1.26$\times$ & 9.99$\times$ \\
    HotpotQA-bridge & +0.66 & +6.41 & 1.25$\times$ & 6.01$\times$ \\
    \bottomrule
  \end{tabular}
\end{table*}

\begin{table*}[t]
  \centering
  \caption{Ablation summary over 27 fully matched cells on QASC, ROPES, and StrategyQA, recomputed from the displayed detailed entries. Tied best values are counted for each tied variant.}
  \label{tab:ablation-summary}
  \begin{tabular}{lrrrrrr}
    \toprule
    Variant & $\Delta$ score & Avg. rank & Best/tied-best & Component (ms) & E2E (ms) & Params \\
    \midrule
    Full & +0.00 & 1.87 & 13/27 & 5.8 & 131.7 & 4.55M \\
    $-\mathrm{RP}$ & -0.46 & 2.35 & 5/27 & 4.7 & 128.7 & 3.58M \\
    $-\mathrm{RX}$ & -0.83 & 3.02 & 5/27 & 3.7 & 127.1 & 3.43M \\
    LCFP & -0.94 & 2.76 & 5/27 & 39.1 & 183.7 & 3.53M \\
    \midrule
    $-\mathrm{RP}$ vs. Full & -0.46 & -- & -- & -18.5\% & -2.3\% & -21.2\% \\
    $-\mathrm{RX}$ vs. Full & -0.83 & -- & -- & -36.9\% & -3.5\% & -24.5\% \\
    LCFP vs. Full & -0.94 & -- & -- & +572.6\% & +39.4\% & -22.3\% \\
    \bottomrule
  \end{tabular}
\end{table*}

\end{document}